\documentclass[11pt,a4paper]{article}
\usepackage{times,latexsym}
\usepackage{url}
\usepackage[T1]{fontenc}
\usepackage{pifont}

\usepackage[acceptedWithA]{tacl2021v1}

\usepackage{xspace,mfirstuc,tabulary}

\usepackage{booktabs}
\usepackage{multirow}
\usepackage{graphicx}
\usepackage{amsmath}
\usepackage{caption}
\usepackage{subcaption}
\usepackage{cleveref}
\usepackage[normalem]{ulem}
\usepackage{xcolor}
\usepackage{ragged2e}

\usepackage[frozencache,cachedir=.]{minted}

\definecolor{magicmint}{rgb}{0.67, 0.94, 0.82}
\definecolor{Box3Color}{RGB}{255, 214, 193}
\PassOptionsToPackage{dvipsnames}{xcolor}

{}

\newcommand{\Dataset}{HAGRID}
\newcommand\Title{{\usefont{T1}{cinzeldecorativebold}{m}{n}{\Dataset}}}
\newcommand\Stylized{{\usefont{T1}{cinzeldecorative}{m}{n}{\Dataset}}}
\newcommand{\miracl}{MIRACL\xspace}

\newif\iftaclinstructions
\taclinstructionsfalse %
\iftaclinstructions
\renewcommand{\confidential}{}
\renewcommand{\anonsubtext}{(No author info supplied here, for consistency with
TACL-submission anonymization requirements)}
\newcommand{\instr}
\fi

\iftaclpubformat %

\else

\fi

\title{{\Title}: A Human-LLM Collaborative Dataset for Generative Information-Seeking with Attribution}

\author{
Ehsan Kamalloo\Thanks{Equal Contribution}$^{*\dagger}$ \quad Aref Jafari$^{*\dagger}$ \quad Xinyu Zhang$^{\dagger}$ \\ 
{\bf Nandan Thakur}$^{\dagger}$ \quad
 {\bf Jimmy Lin}$^{\dagger}$ \\[1ex]
  $^{\dagger}$ David R. Cheriton School of Computer Science, University of Waterloo \\[1ex]
  \texttt{ekamalloo@uwaterloo.ca} \\
}

\date{}

\begin{document}
\maketitle
\begin{abstract}
The rise of large language models (LLMs) had a transformative impact on search, ushering in a new era of search engines that are capable of generating search results in natural language text, imbued with citations for supporting sources.
Building generative information-seeking models demands openly accessible datasets, which currently remain lacking.
In this paper, we introduce a new dataset, {\Dataset} (\textbf{H}uman-in-the-loop \textbf{A}ttributable \textbf{G}enerative \textbf{R}etrieval for \textbf{I}nformation-seeking \textbf{D}ataset) for building end-to-end generative information-seeking models that are capable of retrieving candidate quotes and generating attributed explanations.
Unlike recent efforts that focus on human evaluation of black-box proprietary search engines, we built our dataset atop the English subset of \miracl, a publicly available information retrieval dataset.
{\Dataset} is constructed based on human and LLM collaboration.
We first automatically collect attributed explanations that follow an in-context citation style using an LLM, i.e. GPT-3.5.
Next, we ask human annotators to evaluate the LLM explanations based on two criteria: informativeness and attributability.
{\Dataset} serves as a catalyst for the development of information-seeking models 
with better attribution capabilities.\footnote{{\Dataset} is released at \url{https://github.com/project-miracl/hagrid}.}
\end{abstract}

\section{Introduction}

\begin{table}[t]
\centering
\resizebox{.5\textwidth}{!}{ 
\begin{tabular}{p{8cm}}
\toprule
\textbf{Question}\\
What was Octavia E. Butler's first novel?\\
\midrule
\textbf{Quotes}\\
{[}1{]} Survivor is a science fiction novel by American writer Octavia E. Butler. First published in 1978 as part of Butler's ``Patternist series''...\\
{[}2{]} Butler's first work published was ``Crossover'' in the 1971 Clarion Workshop anthology... Starting in 1974, Butler worked on a series of novels that would later be collected as the Patternist series...\colorbox{magicmint}{The first novel, ``Patternmaster'' (1976),} eventually became the last installment in the series' internal chronology...\\
\midrule
\textbf{Answer}\\
Octavia E. Butler's first novel was ``Patternmaster'' which was published in 1976 and was also the first installment in her ``Patternist series'' {[}2{]}.\\
\midrule
\textbf{Informative?} Yes\\
\textbf{Attributable?} Yes\\
\bottomrule
\end{tabular}
}
\caption{An example taken from {\Stylized} that includes a question along with a list of relevant passages (quotes), an answer generated by GPT-3.5 (\S\ref{sec:answer-gen}), and informativeness and attributability evaluated by human annotators (\S\ref{sec:human}).}
\label{table:example}
\end{table}

Large Language Models (LLMs) have paved the way for the emergence of generative information-seeking search engines such as Bing Chat, Google Bard, and perplexity.ai, where search results are formulated in natural language text, incorporating references to the relevant web pages from which they are derived.
This approach aims to provide users with contextually rich responses.
Yet, LLMs are known to generate text lacking sufficient grounding to knowledge sources \cite{dziri-etal-2022-origin,ji2023survey}, thereby posing risks of misinformation and even worse, hallucination \cite{maynez-etal-2020-faithfulness,raunak-etal-2021-curious}.
This problem becomes particularly critical within search engines where such inaccuracies can erode user trust and potentially spread misinformation \cite{metzler2021rethinking,shah2022situating}.
Building models that are capable of incorporating citations that link to some supporting evidence is a vital step toward understanding the behaviour of LLMs,
allowing users to easily verify the factuality of model outputs.
The development of such models further fosters interpretable LLMs and attributable outputs \cite{rashkin2023measuring}, thus reinforcing the transparency and reliability of LLMs.

A significant obstacle in building generative search models equipped with citations is the lack of accessible and openly available datasets.
The data used by commercial search engines for training their generative information-seeking models are typically proprietary and not accessible to the public, thereby hindering their widespread use in the open-source community.

In this paper, we introduce a new dataset for generative information-seeking scenarios to address these limitations.
Our dataset is constructed on top of \miracl~\cite{miracl}, an information retrieval dataset that consists of information-seeking questions along with a set of manually labeled relevant passages (quotes).
We collect attributed explanations for each question by eliciting prompts from an LLM, i.e., GPT-3.5~\cite{instructgpt}, based on the given relevant passages. The explanations adhere to an in-context citation style, similar to scientific articles, that references the supporting quotes.
We next ask human annotators to judge the explanations based on two criteria, (i) {\em informativeness}: whether the explanation provides a direct answer to the question, and (ii) {\em attributability}: whether the explanation is attributable to the source passages.
We name our dataset {\Dataset}, representing \textbf{H}uman-in-the-loop \textbf{A}ttributable \textbf{G}enerative \textbf{R}etrieval for \textbf{I}nformation-seeking \textbf{D}ataset.
An example question along with its relevant passage and the generated answer is presented in Table~\ref{table:example}.

{\Dataset} consists of two subsets: training and development, enabling researchers to train and evaluate future information-seeking models with attribution capabilities.
In particular, we seek to establish a dataset for building open-source end-to-end search models capable of retrieving candidate quotes and generating attributable answers based on input queries, which are key ingredients in retrieval-augmented generative models~\cite{lewis2020retrieval,izacard-grave-2021-leveraging,borgeaud2022improving}.
In contrast to existing datasets \cite{liu2023evaluating,gao2023enabling}, our emphasis on both openness and the integration of human annotations makes {\Dataset} a valuable and unique resource in this area.
{\Dataset} is publicly released under the Apache 2.0 License.
We hope that open-sourcing of the dataset spurs innovation and further advancements in the rapidly growing area of generative search.

\begin{figure*}[t]
  \centering
  \includegraphics[width=\linewidth]{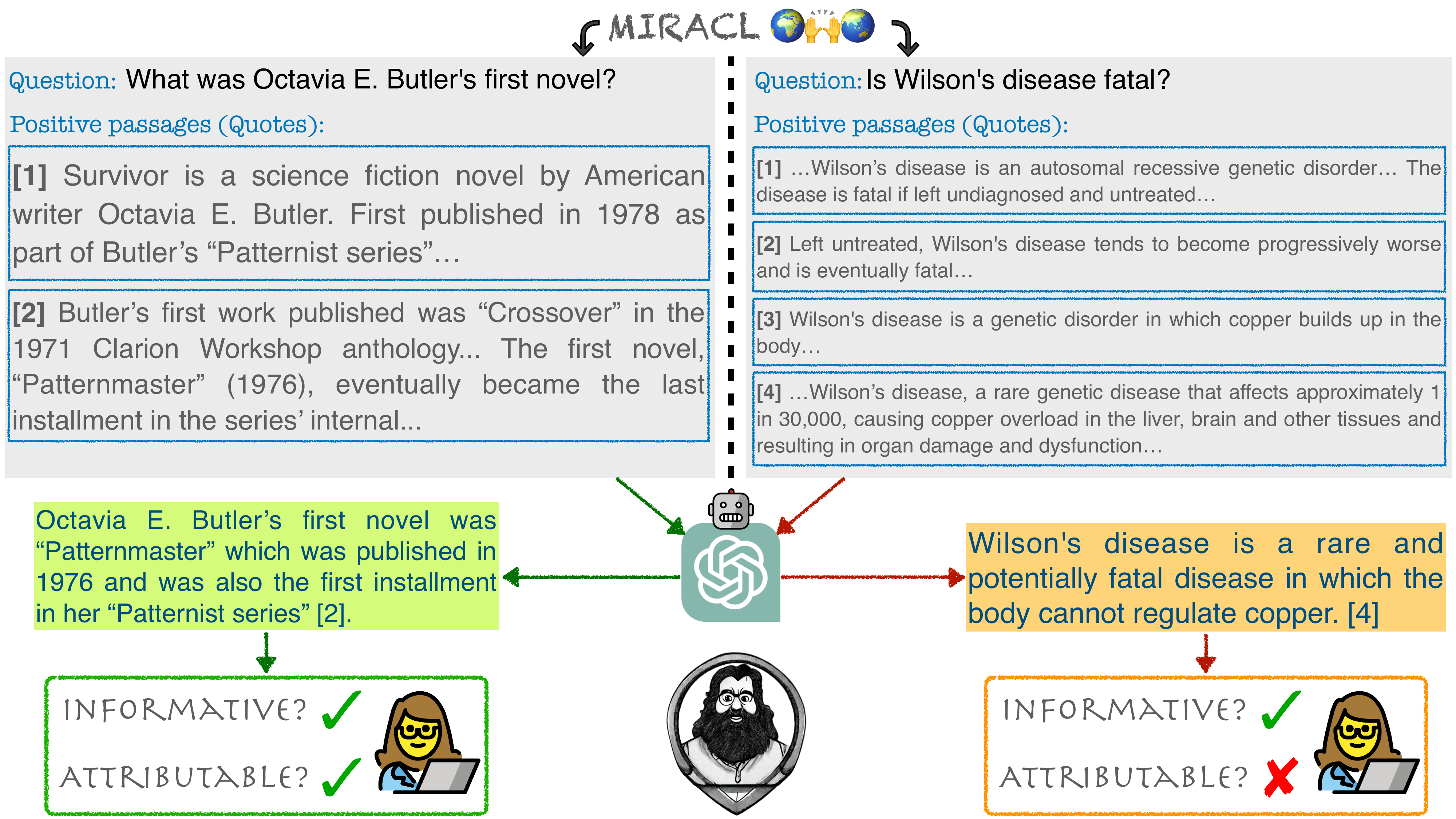}
  \caption{{\Dataset}'s data collection workflow.
  We first take a question and positive passages (quotes) from \miracl, and reformat them into a prompt to instruct an LLM to generate answers with in-context citations (See an example in \autoref{fig:prompt}).
  The answers generated by the LLM are evaluated by human annotators based on two criteria: {\em informativeness} (whether they correctly fulfill the question), and {\em attributability} (whether the source quotes appearing in the answers support the answers). }
  \label{fig:annotation}
\end{figure*}

\section{Related Work}
\paragraph{Explainability.}
Understanding why models behave in certain ways is crucial in deploying them in real-world applications \cite{doshi2017towards}.
A common approach for explainability in NLP is to provide human-understandable explanations for particular outputs of a black-box model \cite{camburu2018snli}.
Numerous attempts were made in many language understanding tasks including text classification \cite{camburu2018snli,liu-etal-2019-towards-explainable}, question answering \cite{abujabal-etal-2017-quint,rajani-etal-2019-explain}, fact verification \cite{atanasova-etal-2020-generating-fact,kotonya-toni-2020-explainable-automated}, and summarization \cite{li2021ease} to generate rationales that explain models' outputs. While these explanations are in line with our goal in this paper, they are not necessarily attributable~\cite{jacovi-goldberg-2020-towards}.
Moreover, several benchmarks \cite{deyoung-etal-2020-eraser,mathew2021hatexplain} were proposed to evaluate the generated rationales.
Towards this goal, \citet{narang2020wt5} built a general-purpose T5 model that generates explanations for its predictions.

\paragraph{Attributability.} \citet{rashkin2023measuring} formalize an attributable statement to identified sources such that it can be entailed from some underlying corpus by a generic hearer. %
Thus, attributability is a specific form of explainability within the constraints of a given source.
WebGPT~\cite{webgpt} and GopherCite~\cite{menick2022teaching} are two recent closed-source models that are capable of generating references to their supporting evidence.
From the data perspective, several QA datasets \cite{geva-etal-2021-aristotle,bohnet2022attributed} provide pointers to text snippets supporting the gold answer.
Moreover, two recent works \cite{liu2023evaluating,gao2023enabling} focus on verifying citations in generated text based on a given set of quotes, which closely aligns with our objective in this paper.
Specifically, \citet{liu2023evaluating} focus on closed-source proprietary search engines, whereas our goal is to use publicly available data to allow for building open-source end-to-end search models.
Similarly, ALCE~\cite{gao2023enabling}, a concurrent work to {\Dataset}, shares a similar goal, albeit with two notable differences.
First, \citet{gao2023enabling} derive questions from QA datasets that consist of question and answer pairs but lack annotations for quotes.
To identify quotes, a retrieval model is adopted to determine relevant passages by matching them with gold answers.
However, automated answer equivalence is shown to be fallible, especially for long-form answers \cite{kamalloo-etal-2023-evaluating,xu-etal-2023-critical} that could lead to accepting irrelevant quotes or rejecting legitimate quotes.
Second, ALCE automatically determines the generated answer correctness as well as the citation quality, in contrast with {\Dataset} where we employ human annotators to validate these important criteria, thereby minimizing the risk of propagating any tool errors into the dataset.

\paragraph{Using LLMs for Dataset Creation.}
Due to the substantial costs, time constraints, and potential biases associated with human data collection, researchers sometimes resort to leveraging machines to reduce the human involvement.
With the advent of proficient LLMs, machine-assisted techniques have become viable to some degree \cite{saunders2022self,wiegreffe-etal-2022-reframing} and in various downstream tasks including natural language inference~\cite{liu-etal-2022-wanli}, instruction-following~\cite{wang-etal-2023-self-instruct,honovich-etal-2023-unnatural}, and information retrieval~\cite{inpars,jeronymo2023inpars}, data collection has evolved into a collaborative effort between models and humans.

\section{Data Collection}

\subsection{Task Formulation}
We characterize the task of attributable information-seeking as the following: 
given a query $Q$ and $n$ text snippets $\mathcal{S}= s_1, \cdots, s_n$ that are relevant to $Q$, the goal is to formulate an answer $A$ to $Q$ such that statements in $A$ are cited to their supporting source snippets $s_i$ based on which they are generated. 
Specifically, answer $A$ is composed of $m$ sentences $a_1, \cdots, a_m$; each ends with a reference $[r_{a_j}]$ where $r_{a_j}$ is a set of integers referring to the indexes of snippets in $\mathcal{S}$;
that is, $r_{a_j} \in \{1.. n\}$.
Note that although certain cases such as ``{\em according to} [1]...'' or ``...{\em supply chain} [2]'' are not explicitly covered by this formulation, such sentences can often be rewritten to follow the specified format.
Also, generic sentences like ``{\em Below is an explanation.}'' do not require citation, but they are not common in information-seeking scenarios.
The examples of cited answers are shown in Figure~\ref{fig:annotation}. 
Our objective is to curate contextualized summary answers derived from a list of text snippets, while also providing the corresponding snippets from which the answers originate.

\subsection{Datasets}
Numerous datasets have been designed for information-seeking scenarios in open-domain QA (\citealt{joshi-etal-2017-triviaqa,lee-etal-2019-latent}; {\em inter alia}) and information retrieval (\citealt{msmarco,soboroff2019trec,voorhees2021trec}; {\em inter alia}). 
In this work, we opt to equip existing retrieval datasets with attribution rather than constructing a dataset from scratch. 
This is primarily motivated by the fact that existing retrieval datasets already contain high-quality queries with judged text snippets, but they typically lack the rationales for annotated answers. 
By leveraging existing queries, we streamline the data collection process, enabling us to focus on attributability. 

\textbf{\miracl}~\cite{miracl}, a multilingual information retrieval (IR) dataset containing queries over Wikipedia articles for 18 diverse languages. 
The evaluated retrieval task setting is monolingual, i.e., both the query and document are of the same language.
The dataset was created using human annotators following a setup similar to \textsc{TyDi} QA \cite{clark-etal-2020-tydi}.
Unlike prior work that segments Wikipedia articles into fixed 100-word passages \cite{karpukhin-etal-2020-dense,clark-etal-2020-tydi,asai-etal-2021-xor}, \miracl split documents based on natural discourse units using two consecutive newlines. 
The dataset represents a standard ad hoc retrieval task, where passages have been marked relevant for each query.
In this work, we focus on working using the English subset of \miracl and leave out other languages for future work.
There are 32.8M passages, 2,863 queries in the training set, and 799 queries in the development set of the \miracl English subset.

\subsection{Answer Generation}\label{sec:answer-gen}
In contrast to QA datasets such as SQuAD~\cite{rajpurkar-etal-2016-squad}, Natural Questions~\cite{kwiatkowski-etal-2019-natural}, or ELI5~\cite{fan-etal-2019-eli5}, questions in \miracl do not have gold answers.
While gold answers could be obtained via human annotations, the effort would be costly and prohibitively time-consuming.
Instead, in our work, we use an existing off-the-shelf LLM to elicit answers because of their ability to effectively generate explanatory answers~\cite{wiegreffe-etal-2022-reframing}.

We input all the positive passages for each query (with at least one relevant passage) in the \miracl dataset into an LLM.
This setup is inspired by retrieval-augmented generation~\cite{lewis2020retrieval}, wherein generation is conditioned not only on the query but also on the retrieved passages.
The relevant passages, derived from the English Wikipedia in MIRACL, will be referred to as ``Quotes.''
As reported in \autoref{table:stats}, nearly 3 quotes on average are provided for each query.
We instruct GPT-3.5, i.e., \texttt{gpt-3.5-turbo-0301}~\cite{chatgpt}, to generate an answer to a question in a zero-shot fashion.
We do not prepare any demonstrations or instructions for prompting with GPT-3.5.
We provide an instruction, and a list of quotes as contexts and ask the LLM to reference answers within brackets \texttt{[]} in the IEEE format.
The complete instruction used is provided in Figure~\ref{fig:prompt}.
We also explored several instructions in the prompt to guide the LLM in generating both short and long answers, leading us to collect multiple answers per query.
However, we found no significant differences between these generated answers.
All the quotes can easily fit within the GPT-3.5 context window size of 4,096 tokens.

We further post-processed model responses to verify the format of model responses using regular expressions and filtered out the ones that violate the specified format. 

\subsection{Human Annotation}
\label{sec:human}
For human assessment, we hired 4 specialist annotators with 1+ year of experience with text data annotation on our team. 
Each annotator was interviewed prior to being hired and was verified to be a fluent and efficient annotator.
To minimize any potential biases and ensure consistency in the annotation process, our team implemented a carefully designed onboarding procedure with training sessions specifically tailored to this task.
The annotators were remunerated with an hourly rate of \$15.2 USD.
In total, the project required approximately 1,400 annotation hours to complete.

Before proceeding with answer annotation, we initially decomposed answers into sentences.
This is in large part to simplify the task as individual sentences are easier to read and evaluate, thus accelerating the data annotation process.
It also allows for collecting fine-grained annotations.
If a sentence lacks citations, we group it with the following sentence that includes a citation, as the citation may pertain to all the grouped sentences.
Following this pre-processing step, we asked our human evaluators to assess two criteria in generated responses:
\begin{itemize}
    \item \textbf{Informativeness} checks whether a generated answer provides a useful response to the question.
    More precisely, if at least one sentence within an answer is labelled informative, the entire answer is deemed informative.
    In essence, this criterion is identical to {\em perceived utility} in \citet{liu2023evaluating}.
    Notably, informativeness encompasses a broader scope, compared to {\em correctness} in \citet{gao2023enabling,liu2023webglm} because it ensures accuracy as well as relevance by taking additional information in the answers into account.
    
    \item \textbf{Attributability} measures whether factual claims in a generated answer can be supported by corresponding quotes.
    An answer sentence would be labelled attributable only if it is fully supported by a cited quote.
    In cases where multiple citations appear in an answer sentence, all cited quotes must contain ample evidence to validate the sentence.
    When all sentences within an answer are labelled attributable, the answer is deemed attributable.
    We observed that annotating attributability takes 3-5x longer than annotating informativeness since annotators should carefully read all cited quotes to arrive at a decision.
    This is why, we were not able to collect annotations for all generated answers due to budget constraints.
\end{itemize}

\subsection{Statistics}
\autoref{table:stats} provides an overview of {\Dataset} in the answer generation phase, prior to human annotation.
The training and development sets contain 1,922 and 716 questions, respectively. 
Using GPT-3.5, we generate around 3,214 (1.7 per question on average) and 1,318 (1.8 on average) answers for train and development sets accordingly.
Moreover, 6,577 and 3,305 citations (2.0 and 2.5 per answer on average) were generated within answers.

The statistics of the annotation results are reported in \autoref{table:labels}.
All the generated answers have been manually evaluated for informativeness,
while around 24\% (754) and 88\% (1,157) of the answers have been evaluated for attributability on the training and development sets, respectively.
The distributions of both informativeness and attributability are greatly consistent between the training and development sets (Informative: 84\% and 90\% answers marked ``yes''; Attributable: 73\% and 71\% answers marked ``yes'', respectively for training and development sets).

\section{{\Dataset} Analysis}
This section presents an in-depth analysis of the {\Dataset} dataset and discusses our main observations. Our aim is two-fold: (1) examining the content of answers with respect to the two criteria, introduced in \S\ref{sec:human}, and (2) how quotes are cited in answers.

\begin{figure}

\tiny
\justifying
\begin{minted}[fontsize=\footnotesize, frame=lines, frame=single,linenos=false,breaklines,breaksymbol=,escapeinside=||,bgcolor=Box3Color]{text}
I will give a question and several context texts about the question. Based on the given contexts, give a brief answer to the question. Also, mention the reference of parts of your answer based on the given contexts within brackets [] as in the IEEE format.

QUESTION:
What was Octavia E. Butler's first novel?

CONTEXTS:
[1] Survivor is a science fiction novel by American writer Octavia E. Butler. First published in 1978 as part of Butler's "Patternist series"...
[2] Butler's first work published was "Crossover" in the 1971 Clarion Workshop anthology... Starting in 1974, Butler worked on a series of novels that would later be collected as the Patternist series... The first novel, "Patternmaster" (1976), eventually became the last installment in the series' internal chronology...

ANSWER:

\end{minted}

\caption{A sample answer generation prompt template that we used in our work for eliciting answers from GPT-3.5~\cite{chatgpt}.}
\label{fig:prompt}
\end{figure}

\begin{table}[t]
\centering
\resizebox{.48\textwidth}{!}{ 
\begin{tabular}{lcc}
\toprule
& \textbf{Train} & \textbf{Dev} \\
\midrule
\#~Questions & 1,922& 716\\
\ding{229} Avg. \#~Quotes per Question & 2.7& 2.9\\
\hline
\#~Generated Answers & 3,214& 1,318\\
\ding{229} Avg. Answer per Question & 1.7& 1.8\\
\midrule
\#~Citations & 6,577& 3,305\\
\ding{229} Avg. Citation per Answer& 2.0& 2.5\\
\bottomrule
\end{tabular}
}
\caption{
The statistics of the answers generated by LLM, prior to human annotation.
Given a question, the model generates multiple answers ({\em Generated Answers}, and each answer may be accompanied by more than one cited quotes ({\em Citations}). 
}
\label{table:stats}
\end{table}

\begin{table}[t]
\centering
\resizebox{.48\textwidth}{!}{ 
\begin{tabular}{lrrcrc}
\toprule
& \textbf{Total}& \multicolumn{2}{c}{\textbf{Yes}}& \multicolumn{2}{c}{\textbf{No}}\\
\midrule
\multicolumn{6}{c}{\textbf{Train}}\\
Informative& 3,214& 2,704& 84\%& 510& 16\%\\
Attributable& 754& 547& 73\%& 207& 27\%\\
\midrule
\multicolumn{6}{c}{\textbf{Dev}}\\
Informative& 1,318& 1,179& 90\%& 139& 10\%\\
Attributable& 1,157& 826& 71\%& 331& 29\%\\
\bottomrule
\end{tabular}
}
\caption{
The statistics of the final {\Dataset} dataset.
\textbf{Total}: the total number of annotated datapoints for each criterion. 
\textbf{Yes/No}: the number of datapoints that are annotated as ``yes'' or ``no'' for each criterion.
}
\label{table:labels}
\end{table}

\begin{figure*}[t]
     \centering
     \begin{subfigure}[b]{0.42\textwidth}
     \includegraphics[width=\textwidth]{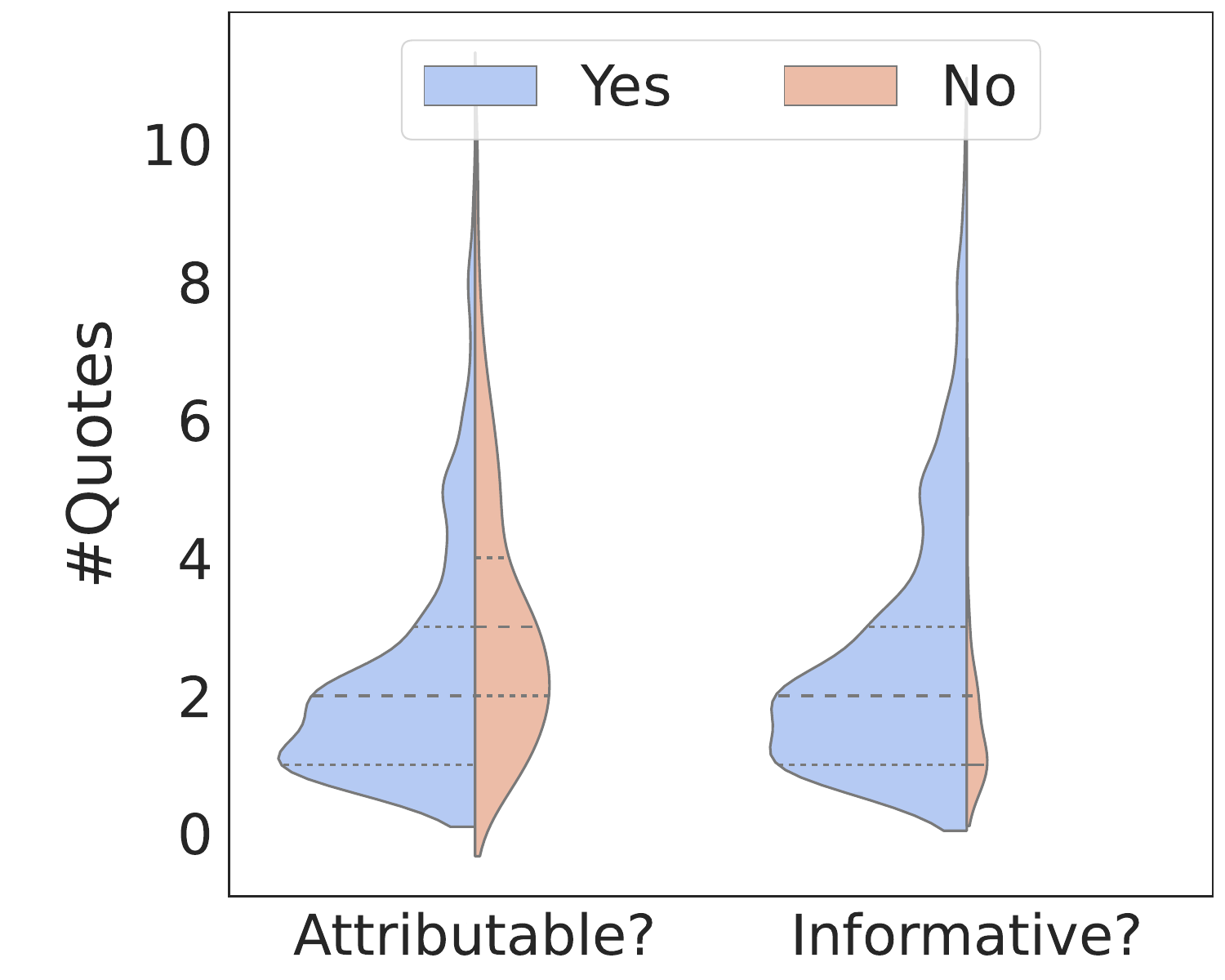}
         \caption{Number of associated quotes}
         \label{fig:num-quotes}
     \end{subfigure}
     \hfill
    \begin{subfigure}[b]{0.42\textwidth}
        \includegraphics[width=\textwidth]{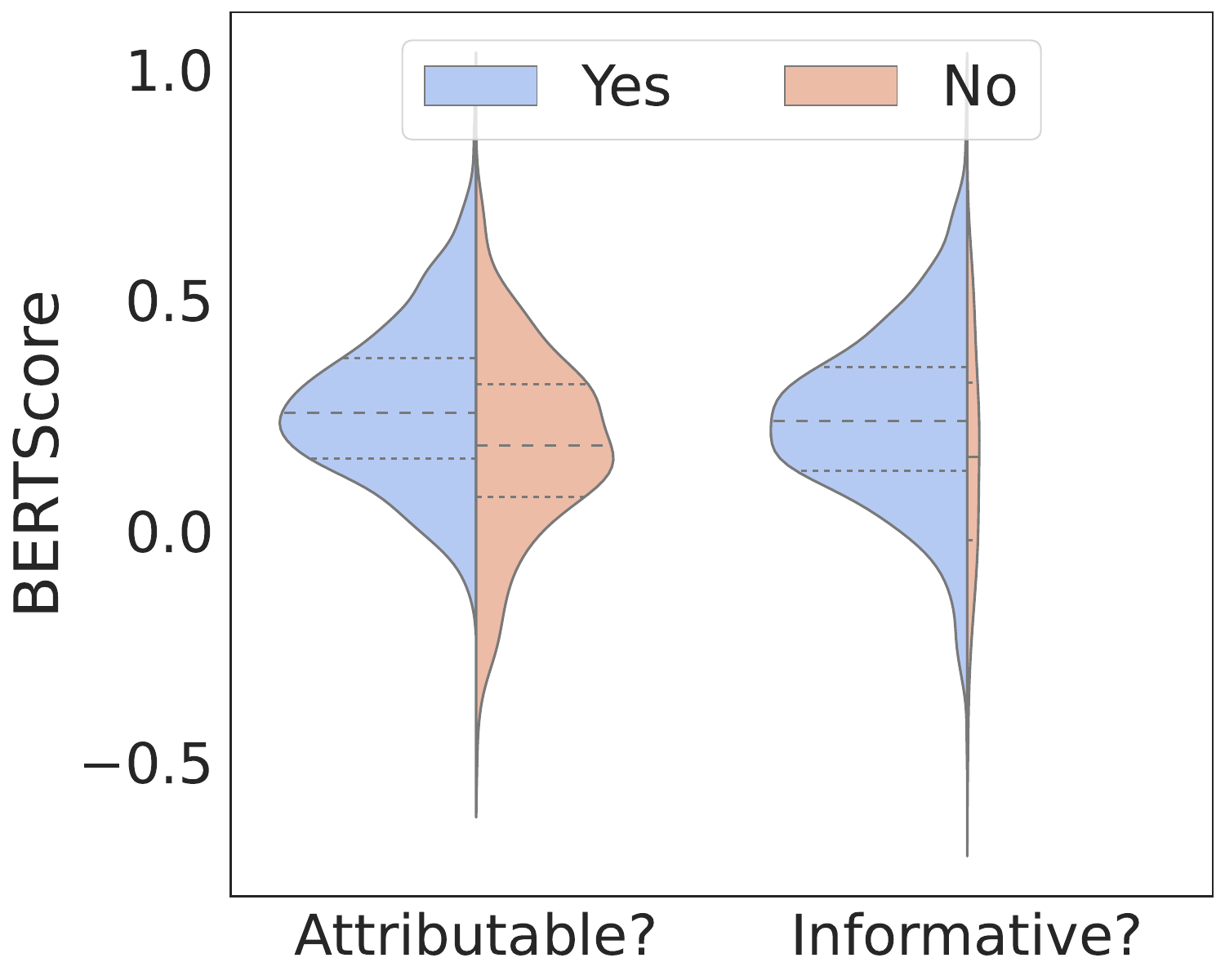}
        \caption{BERTScore}
        \label{fig:bert-score}
    \end{subfigure}
    \caption{Comparative analyses of the y-axis distribution with respect to different answer labels: attributable vs.\ unattributable and informative vs.\ uninformative. Each curve is a KDE plot to represent distribution. The horizontal dash lines depict the first, second, and third quartiles.}
    \label{fig:analysis}
\end{figure*}

\paragraph{Unattributable and uninformative answers often correspond to higher number of quotes.} Figure~\ref{fig:num-quotes} depicts the impact of the number of associated quotes with respect to informativeness and attributability. The mean distribution of the number of quotes for unattributable and uninformative answers is higher than that of attributable and informative answers. Hence, as the number of associated quotes grows, LLMs tend to become fallible in generating informative and attributable answers.

\paragraph{Attributable answers are semantically close to their referring quotes.} Figure~\ref{fig:bert-score} plots the distribution of semantic similarity, measured using BERTScore~\cite{BERTScore}, between answer sentences and corresponding cited sentences in quotes. Looking at the median of the distribution, informative answers are only slightly more semantically similar to the quotes than uninformative answers. Similarly, the difference between the median of the distributions for attributable and unattributable answers with respect to their corresponding quotes is marginal.

\begin{figure}[t]
  \centering
  \includegraphics[width=\linewidth]{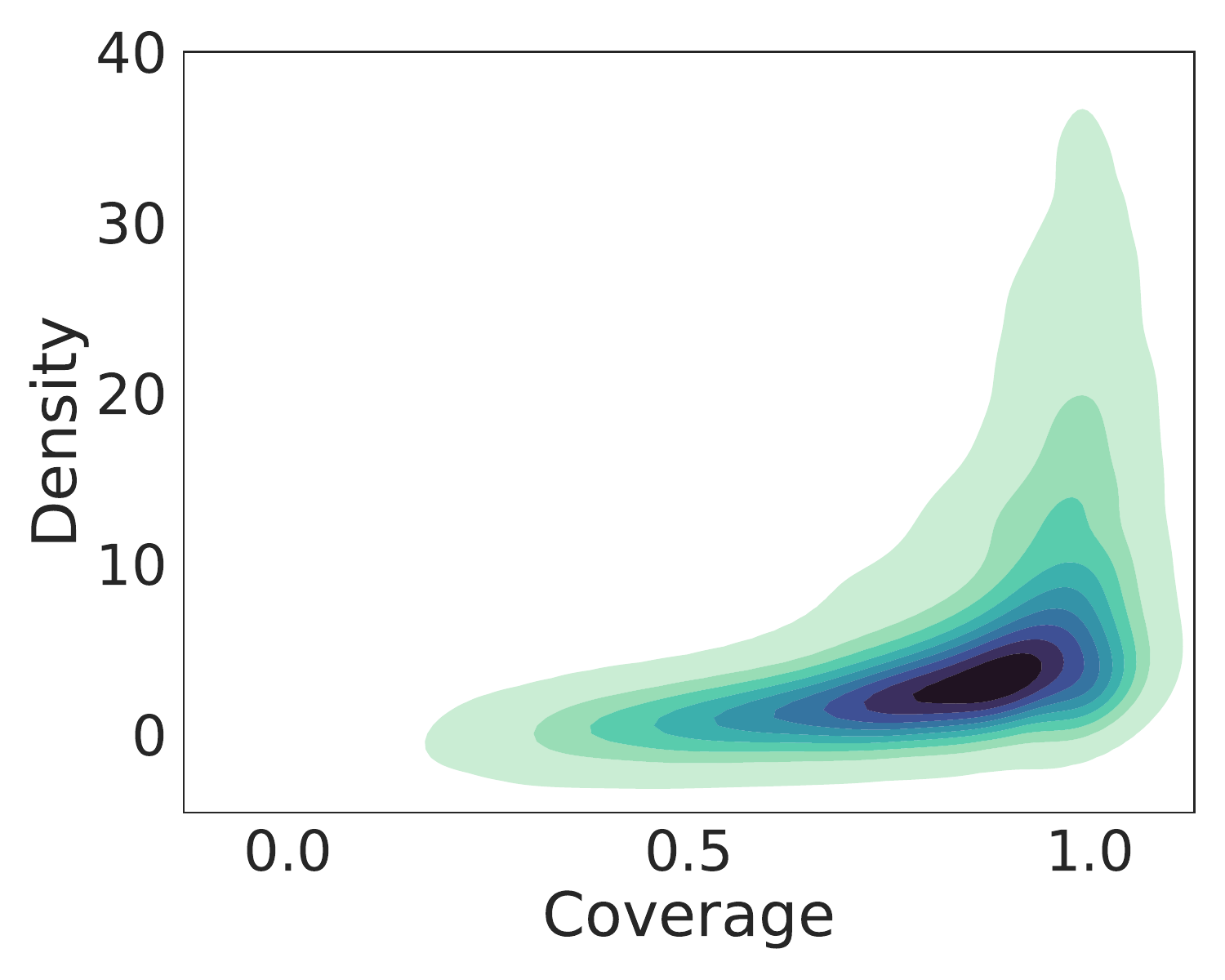}
  \caption{Coverage vs. Density between generated answers and their cited quotes. Answers tend to be extractive, thus words are frequently copied from quotes into the answers.}
  \label{fig:abst-vs-extract}
\end{figure}
\paragraph{Generated answers are often extractive.}
Our goal here is to measure the extent to which generated answers copy text from their corresponding quotes.
To this end, we borrow two metrics, namely coverage and density, from abstractive summarization \cite{grusky-etal-2018-newsroom} whose aims are to gauge extractiveness and abstractiveness in text summaries:
{\em Coverage} that measures the percentage of words in the answer that are also present in the quotes.
A higher coverage signifies greater word overlap between an answer and its corresponding quotes.
And, {\em Density} that quantifies the average length of text fragments from the quotes which subsume answer words.
Answers with larger chunks of text copied from their quotes will result in higher density.
Figure~\ref{fig:abst-vs-extract} illustrates the coverage and density distributions.
While coverage largely falls between 0.5 and 1.0, density is more varied. These results indicate that generated answers tend to use words from their associated quotes and thus, are mostly extractive.

\begin{figure}[t]
  \centering
  \includegraphics[width=\linewidth]{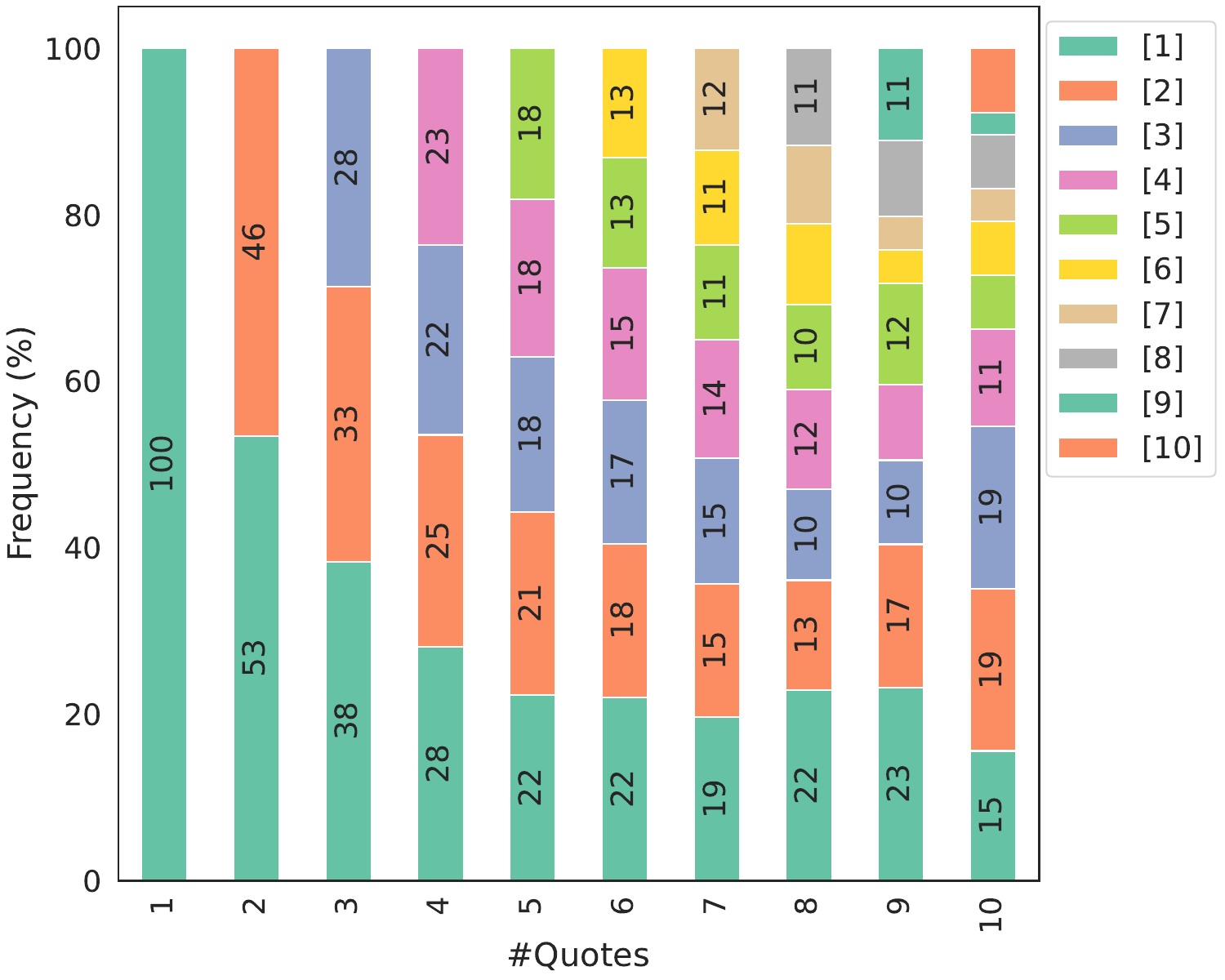}
  \caption{
  Frequency of citation indices based on different numbers of given quotes;
  \textbf{x-axis}: the number of given quotes in the prompt;
  \textbf{y-axis}: the frequency of indices in the citation.
  The citations percentage that are larger than or equal to 10\% are marked on the figure.
  }
  \label{fig:citations}
\end{figure}

\paragraph{When the number of quotes is small, quotes are cited nearly evenly.}
We analyze the citation frequency to determine which quotes are referenced in the generated answers.
The findings are illustrated in Figure~\ref{fig:citations} where the x-axis represents the number of associated quotes and the y-axis shows the percentage of the cited quotes.
When the number of quotes remains below 5, the indices of cited quotes are distributed evenly in general.
However, when the number of quotes surpasses 5, the top-3 quotes receive higher citations, while the lower-ranked quotes are cited far less frequently.

\section{Conclusion}
Generative search with the ability to cite supporting sources has gained a lot of traction lately.
However, the absence of accessible high-quality data inhibits progress in building open-source information-seeking models.
In this paper, we seek to bridge this gap in the community by introducing {\Dataset}, a new dataset for building end-to-end generative retrieval models.
Our dataset is collected via a human-machine collaboration that starts with generating explanatory answers to information-seeking queries from GPT-3.5, followed by a human assessment of correctness and attributability of the generated answers.
{\Dataset} facilitates the development of open-source models for information-seeking scenarios.
Our human study has shed light on the room for improvement, i.e. around 40\% of GPT-3.5 generated answers are not informative and over 20\% fail to demonstrate attribution to the quotes.
Moving forward, future research endeavors may focus on building more accurate models, aimed at mitigating the errors commonly encountered in current LLMs.

\section*{Limitations}
The scope of our dataset is on information-seeking scenarios that mainly inquire about factual statements that usually do not warrant creative or complex reasoning.
Thus, more challenging questions with multi-hop reasoning \cite{yang-etal-2018-hotpotqa}, discrete reasoning \cite{dua-etal-2019-drop}, etc. are not covered in this study.

Another limitation is that {\Dataset} covers only English.
While the source dataset, \miracl, is multilingual and encompasses 18 languages, we leave non-English languages, either high-resource or low-resource, for future work.

\section*{Ethical Statement}
Although we do not foresee any major risk or negative societal impact of our dataset, models constructed on {\Dataset} may inadvertently produce biased outputs due to the reported tendencies of LLMs to generate stereotypes or biases.
Therefore, care must be exercised for responsible deployment of such models in real-world applications.

\iftaclpubformat

\else
\fi

\bibliography{tacl2021,anthology}

\begin{thebibliography}{56}
\expandafter\ifx\csname natexlab\endcsname\relax\def\natexlab#1{#1}\fi

\bibitem[{Abujabal et~al.(2017)Abujabal, Saha~Roy, Yahya, and
  Weikum}]{abujabal-etal-2017-quint}
Abdalghani Abujabal, Rishiraj Saha~Roy, Mohamed Yahya, and Gerhard Weikum.
  2017.
\newblock \href {https://doi.org/10.18653/v1/D17-2011} {{QUINT}: Interpretable
  question answering over knowledge bases}.
\newblock In \emph{Proceedings of the 2017 Conference on Empirical Methods in
  Natural Language Processing: System Demonstrations}, pages 61--66,
  Copenhagen, Denmark. Association for Computational Linguistics.

\bibitem[{Asai et~al.(2021)Asai, Kasai, Clark, Lee, Choi, and
  Hajishirzi}]{asai-etal-2021-xor}
Akari Asai, Jungo Kasai, Jonathan Clark, Kenton Lee, Eunsol Choi, and Hannaneh
  Hajishirzi. 2021.
\newblock \href {https://doi.org/10.18653/v1/2021.naacl-main.46} {{XOR} {QA}:
  Cross-lingual open-retrieval question answering}.
\newblock In \emph{Proceedings of the 2021 Conference of the North American
  Chapter of the Association for Computational Linguistics: Human Language
  Technologies}, pages 547--564, Online. Association for Computational
  Linguistics.

\bibitem[{Atanasova et~al.(2020)Atanasova, Simonsen, Lioma, and
  Augenstein}]{atanasova-etal-2020-generating-fact}
Pepa Atanasova, Jakob~Grue Simonsen, Christina Lioma, and Isabelle Augenstein.
  2020.
\newblock \href {https://doi.org/10.18653/v1/2020.acl-main.656} {Generating
  fact checking explanations}.
\newblock In \emph{Proceedings of the 58th Annual Meeting of the Association
  for Computational Linguistics}, pages 7352--7364, Online. Association for
  Computational Linguistics.

\bibitem[{Bajaj et~al.(2018)Bajaj, Campos, Craswell, Deng, Gao, Liu, Majumder,
  McNamara, Mitra, Nguyen, Rosenberg, Song, Stoica, Tiwary, and Wang}]{msmarco}
Payal Bajaj, Daniel Campos, Nick Craswell, Li~Deng, Jianfeng Gao, Xiaodong Liu,
  Rangan Majumder, Andrew McNamara, Bhaskar Mitra, Tri Nguyen, Mir Rosenberg,
  Xia Song, Alina Stoica, Saurabh Tiwary, and Tong Wang. 2018.
\newblock \href {https://doi.org/10.48550/arXiv.1611.09268} {{MS} {MARCO}: {A
  Human Generated MAchine Reading COmprehension Dataset}}.
\newblock \emph{arXiv preprint arXiv:1611.09268}.

\bibitem[{{Bohnet} et~al.(2022){Bohnet}, {Tran}, {Verga}, {Aharoni}, {Andor},
  {Baldini Soares}, {Ciaramita}, {Eisenstein}, {Ganchev}, {Herzig}, {Hui},
  {Kwiatkowski}, {Ma}, {Ni}, {Sestorain Saralegui}, {Schuster}, {Cohen},
  {Collins}, {Das}, {Metzler}, {Petrov}, and {Webster}}]{bohnet2022attributed}
Bernd {Bohnet}, Vinh~Q. {Tran}, Pat {Verga}, Roee {Aharoni}, Daniel {Andor},
  Livio {Baldini Soares}, Massimiliano {Ciaramita}, Jacob {Eisenstein}, Kuzman
  {Ganchev}, Jonathan {Herzig}, Kai {Hui}, Tom {Kwiatkowski}, Ji~{Ma}, Jianmo
  {Ni}, Lierni {Sestorain Saralegui}, Tal {Schuster}, William~W. {Cohen},
  Michael {Collins}, Dipanjan {Das}, Donald {Metzler}, Slav {Petrov}, and
  Kellie {Webster}. 2022.
\newblock \href {https://doi.org/10.48550/arXiv.2212.08037} {Attributed
  question answering: Evaluation and modeling for attributed large language
  models}.
\newblock \emph{arXiv preprint arXiv:2212.08037}.

\bibitem[{Bonifacio et~al.(2022)Bonifacio, Abonizio, Fadaee, and
  Nogueira}]{inpars}
Luiz Bonifacio, Hugo Abonizio, Marzieh Fadaee, and Rodrigo Nogueira. 2022.
\newblock \href {https://doi.org/10.1145/3477495.3531863} {{InPars}:
  Unsupervised dataset generation for information retrieval}.
\newblock In \emph{Proceedings of the 45th International ACM SIGIR Conference
  on Research and Development in Information Retrieval}, SIGIR '22, pages
  2387--2392. Association for Computing Machinery.

\bibitem[{Borgeaud et~al.(2022)Borgeaud, Mensch, Hoffmann, Cai, Rutherford,
  Millican, Van Den~Driessche, Lespiau, Damoc, Clark, De~Las~Casas, Guy,
  Menick, Ring, Hennigan, Huang, Maggiore, Jones, Cassirer, Brock, Paganini,
  Irving, Vinyals, Osindero, Simonyan, Rae, Elsen, and
  Sifre}]{borgeaud2022improving}
Sebastian Borgeaud, Arthur Mensch, Jordan Hoffmann, Trevor Cai, Eliza
  Rutherford, Katie Millican, George~Bm Van Den~Driessche, Jean-Baptiste
  Lespiau, Bogdan Damoc, Aidan Clark, Diego De~Las~Casas, Aurelia Guy, Jacob
  Menick, Roman Ring, Tom Hennigan, Saffron Huang, Loren Maggiore, Chris Jones,
  Albin Cassirer, Andy Brock, Michela Paganini, Geoffrey Irving, Oriol Vinyals,
  Simon Osindero, Karen Simonyan, Jack Rae, Erich Elsen, and Laurent Sifre.
  2022.
\newblock Improving language models by retrieving from trillions of tokens.
\newblock In \emph{Proceedings of the 39th International Conference on Machine
  Learning}, volume 162 of \emph{Proceedings of Machine Learning Research},
  pages 2206--2240. PMLR.

\bibitem[{Camburu et~al.(2018)Camburu, Rockt\"{a}schel, Lukasiewicz, and
  Blunsom}]{camburu2018snli}
Oana-Maria Camburu, Tim Rockt\"{a}schel, Thomas Lukasiewicz, and Phil Blunsom.
  2018.
\newblock \href
  {https://proceedings.neurips.cc/paper_files/paper/2018/file/4c7a167bb329bd92580a99ce422d6fa6-Paper.pdf}
  {{e-SNLI}: Natural language inference with natural language explanations}.
\newblock In \emph{Advances in Neural Information Processing Systems},
  volume~31. Curran Associates, Inc.

\bibitem[{Clark et~al.(2020)Clark, Choi, Collins, Garrette, Kwiatkowski,
  Nikolaev, and Palomaki}]{clark-etal-2020-tydi}
Jonathan~H. Clark, Eunsol Choi, Michael Collins, Dan Garrette, Tom Kwiatkowski,
  Vitaly Nikolaev, and Jennimaria Palomaki. 2020.
\newblock \href {https://doi.org/10.1162/tacl_a_00317} {{T}y{D}i {QA}: A
  benchmark for information-seeking question answering in typologically diverse
  languages}.
\newblock \emph{Transactions of the Association for Computational Linguistics},
  8:454--470.

\bibitem[{DeYoung et~al.(2020)DeYoung, Jain, Rajani, Lehman, Xiong, Socher, and
  Wallace}]{deyoung-etal-2020-eraser}
Jay DeYoung, Sarthak Jain, Nazneen~Fatema Rajani, Eric Lehman, Caiming Xiong,
  Richard Socher, and Byron~C. Wallace. 2020.
\newblock \href {https://doi.org/10.18653/v1/2020.acl-main.408} {{ERASER}: {A}
  benchmark to evaluate rationalized {NLP} models}.
\newblock In \emph{Proceedings of the 58th Annual Meeting of the Association
  for Computational Linguistics}, pages 4443--4458, Online. Association for
  Computational Linguistics.

\bibitem[{Doshi-Velez and Kim(2017)}]{doshi2017towards}
Finale Doshi-Velez and Been Kim. 2017.
\newblock \href {https://doi.org/10.48550/arXiv.1702.08608} {Towards a rigorous
  science of interpretable machine learning}.
\newblock \emph{arXiv preprint arXiv:1702.08608}.

\bibitem[{Dua et~al.(2019)Dua, Wang, Dasigi, Stanovsky, Singh, and
  Gardner}]{dua-etal-2019-drop}
Dheeru Dua, Yizhong Wang, Pradeep Dasigi, Gabriel Stanovsky, Sameer Singh, and
  Matt Gardner. 2019.
\newblock \href {https://doi.org/10.18653/v1/N19-1246} {{DROP}: A reading
  comprehension benchmark requiring discrete reasoning over paragraphs}.
\newblock In \emph{Proceedings of the 2019 Conference of the North {A}merican
  Chapter of the Association for Computational Linguistics: Human Language
  Technologies, Volume 1 (Long and Short Papers)}, pages 2368--2378,
  Minneapolis, Minnesota. Association for Computational Linguistics.

\bibitem[{Dziri et~al.(2022)Dziri, Milton, Yu, Zaiane, and
  Reddy}]{dziri-etal-2022-origin}
Nouha Dziri, Sivan Milton, Mo~Yu, Osmar Zaiane, and Siva Reddy. 2022.
\newblock \href {https://doi.org/10.18653/v1/2022.naacl-main.387} {On the
  origin of hallucinations in conversational models: Is it the datasets or the
  models?}
\newblock In \emph{Proceedings of the 2022 Conference of the North American
  Chapter of the Association for Computational Linguistics: Human Language
  Technologies}, pages 5271--5285, Seattle, United States. Association for
  Computational Linguistics.

\bibitem[{Fan et~al.(2019)Fan, Jernite, Perez, Grangier, Weston, and
  Auli}]{fan-etal-2019-eli5}
Angela Fan, Yacine Jernite, Ethan Perez, David Grangier, Jason Weston, and
  Michael Auli. 2019.
\newblock \href {https://doi.org/10.18653/v1/P19-1346} {{ELI}5: Long form
  question answering}.
\newblock In \emph{Proceedings of the 57th Annual Meeting of the Association
  for Computational Linguistics}, pages 3558--3567, Florence, Italy.
  Association for Computational Linguistics.

\bibitem[{Gao et~al.(2023)Gao, Yen, Yu, and Chen}]{gao2023enabling}
Tianyu Gao, Howard Yen, Jiatong Yu, and Danqi Chen. 2023.
\newblock \href {https://doi.org/10.48550/arXiv.2305.14627} {Enabling large
  language models to generate text with citations}.
\newblock \emph{arXiv preprint arXiv:2305.14627}.

\bibitem[{Geva et~al.(2021)Geva, Khashabi, Segal, Khot, Roth, and
  Berant}]{geva-etal-2021-aristotle}
Mor Geva, Daniel Khashabi, Elad Segal, Tushar Khot, Dan Roth, and Jonathan
  Berant. 2021.
\newblock \href {https://doi.org/10.1162/tacl_a_00370} {Did aristotle use a
  laptop? a question answering benchmark with implicit reasoning strategies}.
\newblock \emph{Transactions of the Association for Computational Linguistics},
  9:346--361.

\bibitem[{Grusky et~al.(2018)Grusky, Naaman, and
  Artzi}]{grusky-etal-2018-newsroom}
Max Grusky, Mor Naaman, and Yoav Artzi. 2018.
\newblock \href {https://doi.org/10.18653/v1/N18-1065} {{N}ewsroom: A dataset
  of 1.3 million summaries with diverse extractive strategies}.
\newblock In \emph{Proceedings of the 2018 Conference of the North {A}merican
  Chapter of the Association for Computational Linguistics: Human Language
  Technologies, Volume 1 (Long Papers)}, pages 708--719, New Orleans,
  Louisiana. Association for Computational Linguistics.

\bibitem[{Honovich et~al.(2023)Honovich, Scialom, Levy, and
  Schick}]{honovich-etal-2023-unnatural}
Or~Honovich, Thomas Scialom, Omer Levy, and Timo Schick. 2023.
\newblock \href {https://aclanthology.org/2023.acl-long.806} {Unnatural
  instructions: Tuning language models with (almost) no human labor}.
\newblock In \emph{Proceedings of the 61st Annual Meeting of the Association
  for Computational Linguistics (Volume 1: Long Papers)}, pages 14409--14428,
  Toronto, Canada. Association for Computational Linguistics.

\bibitem[{Izacard and Grave(2021)}]{izacard-grave-2021-leveraging}
Gautier Izacard and Edouard Grave. 2021.
\newblock \href {https://doi.org/10.18653/v1/2021.eacl-main.74} {Leveraging
  passage retrieval with generative models for open domain question answering}.
\newblock In \emph{Proceedings of the 16th Conference of the European Chapter
  of the Association for Computational Linguistics: Main Volume}, pages
  874--880, Online. Association for Computational Linguistics.

\bibitem[{Jacovi and Goldberg(2020)}]{jacovi-goldberg-2020-towards}
Alon Jacovi and Yoav Goldberg. 2020.
\newblock \href {https://doi.org/10.18653/v1/2020.acl-main.386} {Towards
  faithfully interpretable {NLP} systems: How should we define and evaluate
  faithfulness?}
\newblock In \emph{Proceedings of the 58th Annual Meeting of the Association
  for Computational Linguistics}, pages 4198--4205, Online. Association for
  Computational Linguistics.

\bibitem[{Jeronymo et~al.(2023)Jeronymo, Bonifacio, Abonizio, Fadaee, Lotufo,
  Zavrel, and Nogueira}]{jeronymo2023inpars}
Vitor Jeronymo, Luiz Bonifacio, Hugo Abonizio, Marzieh Fadaee, Roberto Lotufo,
  Jakub Zavrel, and Rodrigo Nogueira. 2023.
\newblock \href {https://doi.org/10.48550/arXiv.2301.01820} {{InPars-v2}: Large
  language models as efficient dataset generators for information retrieval}.
\newblock \emph{arXiv preprint arXiv:2301.01820}.

\bibitem[{Ji et~al.(2023)Ji, Lee, Frieske, Yu, Su, Xu, Ishii, Bang, Madotto,
  and Fung}]{ji2023survey}
Ziwei Ji, Nayeon Lee, Rita Frieske, Tiezheng Yu, Dan Su, Yan Xu, Etsuko Ishii,
  Ye~Jin Bang, Andrea Madotto, and Pascale Fung. 2023.
\newblock \href {https://doi.org/10.1145/3571730} {Survey of hallucination in
  natural language generation}.
\newblock \emph{ACM Computing Surveys}, 55(12):1--38.

\bibitem[{Joshi et~al.(2017)Joshi, Choi, Weld, and
  Zettlemoyer}]{joshi-etal-2017-triviaqa}
Mandar Joshi, Eunsol Choi, Daniel Weld, and Luke Zettlemoyer. 2017.
\newblock \href {https://doi.org/10.18653/v1/P17-1147} {{T}rivia{QA}: A large
  scale distantly supervised challenge dataset for reading comprehension}.
\newblock In \emph{Proceedings of the 55th Annual Meeting of the Association
  for Computational Linguistics (Volume 1: Long Papers)}, pages 1601--1611,
  Vancouver, Canada. Association for Computational Linguistics.

\bibitem[{Kamalloo et~al.(2023)Kamalloo, Dziri, Clarke, and
  Rafiei}]{kamalloo-etal-2023-evaluating}
Ehsan Kamalloo, Nouha Dziri, Charles Clarke, and Davood Rafiei. 2023.
\newblock \href {https://aclanthology.org/2023.acl-long.307} {Evaluating
  open-domain question answering in the era of large language models}.
\newblock In \emph{Proceedings of the 61st Annual Meeting of the Association
  for Computational Linguistics (Volume 1: Long Papers)}, pages 5591--5606,
  Toronto, Canada. Association for Computational Linguistics.

\bibitem[{Karpukhin et~al.(2020)Karpukhin, Oguz, Min, Lewis, Wu, Edunov, Chen,
  and Yih}]{karpukhin-etal-2020-dense}
Vladimir Karpukhin, Barlas Oguz, Sewon Min, Patrick Lewis, Ledell Wu, Sergey
  Edunov, Danqi Chen, and Wen-tau Yih. 2020.
\newblock \href {https://doi.org/10.18653/v1/2020.emnlp-main.550} {Dense
  passage retrieval for open-domain question answering}.
\newblock In \emph{Proceedings of the 2020 Conference on Empirical Methods in
  Natural Language Processing (EMNLP)}, pages 6769--6781, Online. Association
  for Computational Linguistics.

\bibitem[{Kotonya and Toni(2020)}]{kotonya-toni-2020-explainable-automated}
Neema Kotonya and Francesca Toni. 2020.
\newblock \href {https://doi.org/10.18653/v1/2020.emnlp-main.623} {Explainable
  automated fact-checking for public health claims}.
\newblock In \emph{Proceedings of the 2020 Conference on Empirical Methods in
  Natural Language Processing (EMNLP)}, pages 7740--7754, Online. Association
  for Computational Linguistics.

\bibitem[{Kwiatkowski et~al.(2019)Kwiatkowski, Palomaki, Redfield, Collins,
  Parikh, Alberti, Epstein, Polosukhin, Devlin, Lee, Toutanova, Jones, Kelcey,
  Chang, Dai, Uszkoreit, Le, and Petrov}]{kwiatkowski-etal-2019-natural}
Tom Kwiatkowski, Jennimaria Palomaki, Olivia Redfield, Michael Collins, Ankur
  Parikh, Chris Alberti, Danielle Epstein, Illia Polosukhin, Jacob Devlin,
  Kenton Lee, Kristina Toutanova, Llion Jones, Matthew Kelcey, Ming-Wei Chang,
  Andrew~M. Dai, Jakob Uszkoreit, Quoc Le, and Slav Petrov. 2019.
\newblock \href {https://doi.org/10.1162/tacl_a_00276} {Natural questions: A
  benchmark for question answering research}.
\newblock \emph{Transactions of the Association for Computational Linguistics},
  7:452--466.

\bibitem[{Lee et~al.(2019)Lee, Chang, and Toutanova}]{lee-etal-2019-latent}
Kenton Lee, Ming-Wei Chang, and Kristina Toutanova. 2019.
\newblock \href {https://doi.org/10.18653/v1/P19-1612} {Latent retrieval for
  weakly supervised open domain question answering}.
\newblock In \emph{Proceedings of the 57th Annual Meeting of the Association
  for Computational Linguistics}, pages 6086--6096, Florence, Italy.
  Association for Computational Linguistics.

\bibitem[{Lewis et~al.(2020)Lewis, Perez, Piktus, Petroni, Karpukhin, Goyal,
  K\"{u}ttler, Lewis, Yih, Rockt\"{a}schel, Riedel, and
  Kiela}]{lewis2020retrieval}
Patrick Lewis, Ethan Perez, Aleksandra Piktus, Fabio Petroni, Vladimir
  Karpukhin, Naman Goyal, Heinrich K\"{u}ttler, Mike Lewis, Wen-tau Yih, Tim
  Rockt\"{a}schel, Sebastian Riedel, and Douwe Kiela. 2020.
\newblock \href
  {https://proceedings.neurips.cc/paper_files/paper/2020/file/6b493230205f780e1bc26945df7481e5-Paper.pdf}
  {Retrieval-augmented generation for knowledge-intensive {NLP} tasks}.
\newblock In \emph{Advances in Neural Information Processing Systems},
  volume~33, pages 9459--9474. Curran Associates Inc.

\bibitem[{Li et~al.(2021)Li, Einolghozati, Iyer, Paranjape, Mehdad, Gupta, and
  Ghazvininejad}]{li2021ease}
Haoran Li, Arash Einolghozati, Srinivasan Iyer, Bhargavi Paranjape, Yashar
  Mehdad, Sonal Gupta, and Marjan Ghazvininejad. 2021.
\newblock \href {https://doi.org/10.48550/arXiv.2105.06982} {{EASE}:
  Extractive-abstractive summarization with explanations}.
\newblock \emph{arXiv preprint arXiv:2105.06982}.

\bibitem[{Liu et~al.(2022)Liu, Swayamdipta, Smith, and
  Choi}]{liu-etal-2022-wanli}
Alisa Liu, Swabha Swayamdipta, Noah~A. Smith, and Yejin Choi. 2022.
\newblock \href {https://aclanthology.org/2022.findings-emnlp.508} {{WANLI}:
  Worker and {AI} collaboration for natural language inference dataset
  creation}.
\newblock In \emph{Findings of the Association for Computational Linguistics:
  EMNLP 2022}, pages 6826--6847, Abu Dhabi, United Arab Emirates. Association
  for Computational Linguistics.

\bibitem[{Liu et~al.(2019)Liu, Yin, and
  Wang}]{liu-etal-2019-towards-explainable}
Hui Liu, Qingyu Yin, and William~Yang Wang. 2019.
\newblock \href {https://doi.org/10.18653/v1/P19-1560} {Towards explainable
  {NLP}: A generative explanation framework for text classification}.
\newblock In \emph{Proceedings of the 57th Annual Meeting of the Association
  for Computational Linguistics}, pages 5570--5581, Florence, Italy.
  Association for Computational Linguistics.

\bibitem[{Liu et~al.(2023{\natexlab{a}})Liu, Zhang, and
  Liang}]{liu2023evaluating}
Nelson~F Liu, Tianyi Zhang, and Percy Liang. 2023{\natexlab{a}}.
\newblock \href {https://doi.org/10.48550/arXiv.2304.09848} {Evaluating
  verifiability in generative search engines}.
\newblock \emph{arXiv preprint arXiv:2304.09848}.

\bibitem[{Liu et~al.(2023{\natexlab{b}})Liu, Lai, Yu, Xu, Zeng, Du, Zhang,
  Dong, and Tang}]{liu2023webglm}
Xiao Liu, Hanyu Lai, Hao Yu, Yifan Xu, Aohan Zeng, Zhengxiao Du, Peng Zhang,
  Yuxiao Dong, and Jie Tang. 2023{\natexlab{b}}.
\newblock \href {https://doi.org/10.48550/arXiv.2306.07906} {{WebGLM}: Towards
  an efficient web-enhanced question answering system with human preferences}.
\newblock \emph{arXiv preprint arXiv:2306.07906}.

\bibitem[{Mathew et~al.(2021)Mathew, Saha, Yimam, Biemann, Goyal, and
  Mukherjee}]{mathew2021hatexplain}
Binny Mathew, Punyajoy Saha, Seid~Muhie Yimam, Chris Biemann, Pawan Goyal, and
  Animesh Mukherjee. 2021.
\newblock \href {https://doi.org/10.1609/aaai.v35i17.17745} {{HateXplain}: A
  benchmark dataset for explainable hate speech detection}.
\newblock In \emph{Proceedings of the AAAI Conference on Artificial
  Intelligence}, volume~35, pages 14867--14875.

\bibitem[{Maynez et~al.(2020)Maynez, Narayan, Bohnet, and
  McDonald}]{maynez-etal-2020-faithfulness}
Joshua Maynez, Shashi Narayan, Bernd Bohnet, and Ryan McDonald. 2020.
\newblock \href {https://doi.org/10.18653/v1/2020.acl-main.173} {On
  faithfulness and factuality in abstractive summarization}.
\newblock In \emph{Proceedings of the 58th Annual Meeting of the Association
  for Computational Linguistics}, pages 1906--1919, Online. Association for
  Computational Linguistics.

\bibitem[{{Menick} et~al.(2022){Menick}, {Trebacz}, {Mikulik}, {Aslanides},
  {Song}, {Chadwick}, {Glaese}, {Young}, {Campbell-Gillingham}, {Irving}, and
  {McAleese}}]{menick2022teaching}
Jacob {Menick}, Maja {Trebacz}, Vladimir {Mikulik}, John {Aslanides}, Francis
  {Song}, Martin {Chadwick}, Mia {Glaese}, Susannah {Young}, Lucy
  {Campbell-Gillingham}, Geoffrey {Irving}, and Nat {McAleese}. 2022.
\newblock \href {https://doi.org/10.48550/arXiv.2203.11147} {Teaching language
  models to support answers with verified quotes}.
\newblock \emph{arXiv preprint arXiv:2203.11147}.

\bibitem[{Metzler et~al.(2021)Metzler, Tay, Bahri, and
  Najork}]{metzler2021rethinking}
Donald Metzler, Yi~Tay, Dara Bahri, and Marc Najork. 2021.
\newblock \href {https://doi.org/10.1145/3476415.3476428} {Rethinking search:
  making domain experts out of dilettantes}.
\newblock \emph{SIGIR Forum}, 55(1):1--27.

\bibitem[{{Nakano} et~al.(2021){Nakano}, {Hilton}, {Balaji}, {Wu}, {Ouyang},
  {Kim}, {Hesse}, {Jain}, {Kosaraju}, {Saunders}, {Jiang}, {Cobbe}, {Eloundou},
  {Krueger}, {Button}, {Knight}, {Chess}, and {Schulman}}]{webgpt}
Reiichiro {Nakano}, Jacob {Hilton}, Suchir {Balaji}, Jeff {Wu}, Long {Ouyang},
  Christina {Kim}, Christopher {Hesse}, Shantanu {Jain}, Vineet {Kosaraju},
  William {Saunders}, Xu~{Jiang}, Karl {Cobbe}, Tyna {Eloundou}, Gretchen
  {Krueger}, Kevin {Button}, Matthew {Knight}, Benjamin {Chess}, and John
  {Schulman}. 2021.
\newblock \href {https://doi.org/10.48550/arXiv.2112.09332} {{WebGPT}:
  Browser-assisted question-answering with human feedback}.
\newblock \emph{arXiv preprint arXiv:2112.09332}.

\bibitem[{Narang et~al.(2020)Narang, Raffel, Lee, Roberts, Fiedel, and
  Malkan}]{narang2020wt5}
Sharan Narang, Colin Raffel, Katherine Lee, Adam Roberts, Noah Fiedel, and
  Karishma Malkan. 2020.
\newblock \href {https://doi.org/10.48550/arXiv.2004.14546} {{WT5?!} training
  text-to-text models to explain their predictions}.
\newblock \emph{arXiv preprint arXiv:2004.14546}.

\bibitem[{OpenAI(2022)}]{chatgpt}
OpenAI. 2022.
\newblock \href {https://openai.com/blog/chatgpt} {Introducing {ChatGPT}}.

\bibitem[{Ouyang et~al.(2022)Ouyang, Wu, Jiang, Almeida, Wainwright, Mishkin,
  Zhang, Agarwal, Slama, Gray, Schulman, Hilton, Kelton, Miller, Simens,
  Askell, Welinder, Christiano, Leike, and Lowe}]{instructgpt}
Long Ouyang, Jeffrey Wu, Xu~Jiang, Diogo Almeida, Carroll Wainwright, Pamela
  Mishkin, Chong Zhang, Sandhini Agarwal, Katarina Slama, Alex Gray, John
  Schulman, Jacob Hilton, Fraser Kelton, Luke Miller, Maddie Simens, Amanda
  Askell, Peter Welinder, Paul Christiano, Jan Leike, and Ryan Lowe. 2022.
\newblock \href {https://openreview.net/forum?id=TG8KACxEON} {Training language
  models to follow instructions with human feedback}.
\newblock In \emph{Advances in Neural Information Processing Systems}, pages
  27730--27744. Curran Associates, Inc.

\bibitem[{Rajani et~al.(2019)Rajani, McCann, Xiong, and
  Socher}]{rajani-etal-2019-explain}
Nazneen~Fatema Rajani, Bryan McCann, Caiming Xiong, and Richard Socher. 2019.
\newblock \href {https://doi.org/10.18653/v1/P19-1487} {Explain yourself!
  leveraging language models for commonsense reasoning}.
\newblock In \emph{Proceedings of the 57th Annual Meeting of the Association
  for Computational Linguistics}, pages 4932--4942, Florence, Italy.
  Association for Computational Linguistics.

\bibitem[{Rajpurkar et~al.(2016)Rajpurkar, Zhang, Lopyrev, and
  Liang}]{rajpurkar-etal-2016-squad}
Pranav Rajpurkar, Jian Zhang, Konstantin Lopyrev, and Percy Liang. 2016.
\newblock \href {https://doi.org/10.18653/v1/D16-1264} {{SQ}u{AD}: 100,000+
  questions for machine comprehension of text}.
\newblock In \emph{Proceedings of the 2016 Conference on Empirical Methods in
  Natural Language Processing}, pages 2383--2392, Austin, Texas. Association
  for Computational Linguistics.

\bibitem[{Rashkin et~al.(2023)Rashkin, Nikolaev, Lamm, Aroyo, Collins, Das,
  Petrov, Singh~Tomar, Turc, and Reitter}]{rashkin2023measuring}
Hannah Rashkin, Vitaly Nikolaev, Matthew Lamm, Lora Aroyo, Michael Collins,
  Dipanjan Das, Slav Petrov, Gaurav Singh~Tomar, Iulia Turc, and David Reitter.
  2023.
\newblock \href {https://doi.org/10.1162/coli_a_00486} {Measuring attribution
  in natural language generation models}.
\newblock \emph{Computational Linguistics}, pages 1--66.

\bibitem[{Raunak et~al.(2021)Raunak, Menezes, and
  Junczys-Dowmunt}]{raunak-etal-2021-curious}
Vikas Raunak, Arul Menezes, and Marcin Junczys-Dowmunt. 2021.
\newblock \href {https://doi.org/10.18653/v1/2021.naacl-main.92} {The curious
  case of hallucinations in neural machine translation}.
\newblock In \emph{Proceedings of the 2021 Conference of the North American
  Chapter of the Association for Computational Linguistics: Human Language
  Technologies}, pages 1172--1183, Online. Association for Computational
  Linguistics.

\bibitem[{Saunders et~al.(2022)Saunders, Yeh, Wu, Bills, Ouyang, Ward, and
  Leike}]{saunders2022self}
William Saunders, Catherine Yeh, Jeff Wu, Steven Bills, Long Ouyang, Jonathan
  Ward, and Jan Leike. 2022.
\newblock \href {https://doi.org/10.48550/arXiv.2206.05802} {Self-critiquing
  models for assisting human evaluators}.
\newblock \emph{arXiv preprint arXiv:2206.05802}.

\bibitem[{Shah and Bender(2022)}]{shah2022situating}
Chirag Shah and Emily~M. Bender. 2022.
\newblock \href {https://doi.org/10.1145/3498366.3505816} {Situating search}.
\newblock In \emph{Proceedings of the 2022 Conference on Human Information
  Interaction and Retrieval}, CHIIR '22, pages 221--232. Association for
  Computing Machinery.

\bibitem[{Soboroff et~al.(2019)Soboroff, Huang, and Harman}]{soboroff2019trec}
Ian Soboroff, Shudong Huang, and Donna Harman. 2019.
\newblock \href {https://trec.nist.gov/pubs/trec28/papers/OVERVIEW.N.pdf}
  {{TREC} 2019 news track overview}.
\newblock In \emph{TREC}.

\bibitem[{Voorhees et~al.(2021)Voorhees, Alam, Bedrick, Demner-Fushman, Hersh,
  Lo, Roberts, Soboroff, and Wang}]{voorhees2021trec}
Ellen Voorhees, Tasmeer Alam, Steven Bedrick, Dina Demner-Fushman, William~R
  Hersh, Kyle Lo, Kirk Roberts, Ian Soboroff, and Lucy~Lu Wang. 2021.
\newblock \href {https://doi.org/10.1145/3451964.3451965} {{TREC-COVID}:
  Constructing a pandemic information retrieval test collection}.
\newblock \emph{SIGIR Forum}, 54(1):1--12.

\bibitem[{Wang et~al.(2023)Wang, Kordi, Mishra, Liu, Smith, Khashabi, and
  Hajishirzi}]{wang-etal-2023-self-instruct}
Yizhong Wang, Yeganeh Kordi, Swaroop Mishra, Alisa Liu, Noah~A. Smith, Daniel
  Khashabi, and Hannaneh Hajishirzi. 2023.
\newblock \href {https://aclanthology.org/2023.acl-long.754} {Self-instruct:
  Aligning language models with self-generated instructions}.
\newblock In \emph{Proceedings of the 61st Annual Meeting of the Association
  for Computational Linguistics (Volume 1: Long Papers)}, pages 13484--13508,
  Toronto, Canada. Association for Computational Linguistics.

\bibitem[{Wiegreffe et~al.(2022)Wiegreffe, Hessel, Swayamdipta, Riedl, and
  Choi}]{wiegreffe-etal-2022-reframing}
Sarah Wiegreffe, Jack Hessel, Swabha Swayamdipta, Mark Riedl, and Yejin Choi.
  2022.
\newblock \href {https://doi.org/10.18653/v1/2022.naacl-main.47} {Reframing
  human-{AI} collaboration for generating free-text explanations}.
\newblock In \emph{Proceedings of the 2022 Conference of the North American
  Chapter of the Association for Computational Linguistics: Human Language
  Technologies}, pages 632--658, Seattle, United States. Association for
  Computational Linguistics.

\bibitem[{Xu et~al.(2023)Xu, Song, Iyyer, and Choi}]{xu-etal-2023-critical}
Fangyuan Xu, Yixiao Song, Mohit Iyyer, and Eunsol Choi. 2023.
\newblock \href {https://aclanthology.org/2023.acl-long.181} {A critical
  evaluation of evaluations for long-form question answering}.
\newblock In \emph{Proceedings of the 61st Annual Meeting of the Association
  for Computational Linguistics (Volume 1: Long Papers)}, pages 3225--3245,
  Toronto, Canada. Association for Computational Linguistics.

\bibitem[{Yang et~al.(2018)Yang, Qi, Zhang, Bengio, Cohen, Salakhutdinov, and
  Manning}]{yang-etal-2018-hotpotqa}
Zhilin Yang, Peng Qi, Saizheng Zhang, Yoshua Bengio, William Cohen, Ruslan
  Salakhutdinov, and Christopher~D. Manning. 2018.
\newblock \href {https://doi.org/10.18653/v1/D18-1259} {{H}otpot{QA}: A dataset
  for diverse, explainable multi-hop question answering}.
\newblock In \emph{Proceedings of the 2018 Conference on Empirical Methods in
  Natural Language Processing}, pages 2369--2380, Brussels, Belgium.
  Association for Computational Linguistics.

\bibitem[{Zhang et~al.(2020)Zhang, Kishore, Wu, Weinberger, and
  Artzi}]{BERTScore}
Tianyi Zhang, Varsha Kishore, Felix Wu, Kilian~Q. Weinberger, and Yoav Artzi.
  2020.
\newblock \href {https://openreview.net/forum?id=SkeHuCVFDr} {{BERTScore}:
  Evaluating text generation with {BERT}}.
\newblock In \emph{International Conference on Learning Representations}.

\bibitem[{Zhang et~al.(2022)Zhang, Thakur, Ogundepo, Kamalloo, Alfonso-Hermelo,
  Li, Liu, Rezagholizadeh, and Lin}]{miracl}
Xinyu Zhang, Nandan Thakur, Odunayo Ogundepo, Ehsan Kamalloo, David
  Alfonso-Hermelo, Xiaoguang Li, Qun Liu, Mehdi Rezagholizadeh, and Jimmy Lin.
  2022.
\newblock \href {https://doi.org/10.48550/arXiv.2210.09984} {Making a {MIRACL}:
  Multilingual information retrieval across a continuum of languages}.
\newblock \emph{arXiv preprint arXiv:2210.09984}.

\end{thebibliography}
\bibliographystyle{acl_natbib}

\iftaclpubformat

\onecolumn

\appendix

\fi

\end{document}